**Computed Decision Weights and a New Learning Algorithm for Neural Classifiers**

**Eugene Wong**
**University of California at Berkeley**

**Abstract**

In this paper we consider the possibility of computing rather than training the decision layer weights of a neural classifier. Such a possibility arises in two way, from making an appropriate choice of loss function and by solving a problem of constrained optimization. The latter formulation leads to a promising new learning process for pre-decision weights with both simplicity and efficacy.

## Introduction

We represent a general neural classifier in the following block diagram:

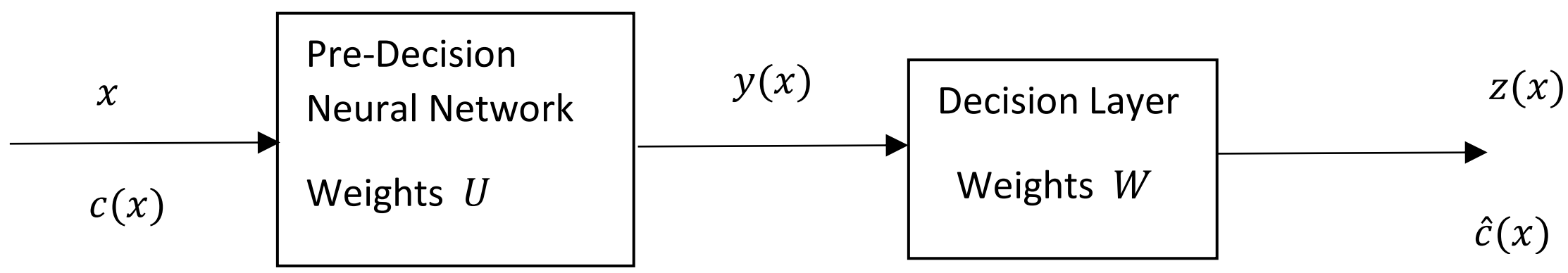

Figure 1. Neural Classifier

The pre-decision network is a general feed-forward neural network with zero or more layers. The input $x$ is an n-vector with real value components. The *class function* $c(x)$ assigns each input to one of $K$ classes. If $c(x)$ is known and used in determining the weights, $x$ is called a *training vector*. If $c(x)$ is known but not used in determining the weights, $x$ is called a *testing vector*.

The decision layer consists of a single layer neural network with a set of weights $w_i$ for each class $i$. The output of the decision layer is a predicted class assignment $\hat{c}(x)$ defined as the class $i$ that maximizes the *decision variables*

$$z_i(x) = w_i \cdot y(x) \tag{1}$$

In batch gradient descent learning algorithms the weights of the entire classifier are trained in a process designed to minimize a loss function $L$ of the form

$$L = \sum_{x \in X} l(x) \tag{2}$$

In (2) $X$ denotes the set of training vectors and $l(x)$ is a function of $\{z_i(x), i \leq K\}$ where $K$ denotes the number of classes.

A commonly used loss function is *cross entropy* [Cy 1999, Zh 2018] defined in terms of estimated posterior probabilities

$$p_i(x) = e^{z_i} / \sum_{j \leq K} e^{z_j} \tag{3}$$

The cross entropy loss function is defined by

$$l(x) = -ln p_{c(x)} \tag{4}$$

Conventional batch learning using gradient descent (GD) proceeds by computing the gradient $\nabla_{w_i} L$ for each class $i$ and adjusts the weights according to the formula

$$\Delta w_i = -\alpha \nabla_{w_i} L \tag{5}$$

where $\alpha$ is a small positive constant commonly known as the learning rate.

In [Wo 2019] we note that the gradient can be expressed as

$$\nabla_{w_i} L = -(M_i - \sum_{x \in X} p_i(x) x) \tag{6}$$

The vectors $M_i$ are defined by

$$M_i = \sum_{x \in X} \delta_{ic(x)} y(x) \tag{7}$$

where $\delta$ denotes the Kronecker delta. We note that $M_i$ are the only entities for which the class labels for the training vectors are used in gradient descent. If, as in [2], we replace $p_i(x)$ by $w_i \cdot y(x)$ in (6) and set the gradient to 0, we then obtain an explicit formula for the weights given as follows:

$$w_i = (YY')^{-1}M_i \qquad (8)$$

Equation (8) is a matrix equation where $w_i$ and $M_i$ are column vectors; $Y$ is a matrix with the vectors $y(x), x \in X$ , as its columns; prime denotes transpose; and $(YY')^{-1}$is the inverse of $YY'$which exists provided that the vectors $\{y(x), x \in X\}$ are linearly independent. In extensive testing the decision weights given by (8) have proven to be surprising good for shallow networks.

Quadratic Loss Function

The cross entropy (4) belongs to a class of loss functions of the form

$$l(x) = -[z_{c(x)} - f(z(x))] \qquad (9)$$

Two simple and interesting examples of this class are:

Cross Entropy: $$f(z(x)) = \ln(\textstyle\sum_{i \le K} e^{z_i(x)}) \qquad (10)$$

Quadratic Loss: $$f\big(z(x)\big) = \tfrac{1}{2}\textstyle\sum_{i \le K} {z_i}^2 \qquad (11)$$

For cross entropy and quadratic loss the gradient formula is given by

$$\nabla_{w_i} l(x) = -\big[\delta_{ic(x)} - p_i(x)\big]y(x) \qquad (12)$$

$$\nabla_{w_i} l(x) = -\big[\delta_{ic(x)} - w_i \cdot y(x)\big]y(x) \qquad (13)$$

An intuitive interpretation of (13) suggests that the quadratic form (11) is a reasonably good loss function for gradient descent. If the weight adjustments are proportional to the negative of the gradient, they increase $z_i = w_i \cdot y(x)$ when $i$ agrees with the class label for $x$ and disproportionally reduce the larger values of $z_i$ that may cause a misclassification of $x$.

The batch gradient of the quadratic loss is found by summing (13) over the training set $X$ and can be expressed in matrix form as follows:

$$\nabla_{w_i} L = -[M_i - (YY')w_i] \tag{14}$$

In (15) $M_i$ and $w_i$ are column vectors and $Y$ is a matrix with columns given by $y(x)$ and indexed by $x$. Prime denotes transpose. It follows that the formula (8) for computing $w_i$ is simply an equilibrium point of (15).

An intuitive interpretation of (13) together with (14) provide a measure of explanation as to why (8) works well on some data. The following data set is used in all our empirical computations:

| | |
|---|---|
| Origin of data: | Pre-processed CIFAR-10 |
| Number of Classes | 10 |
| Dimension of data vector: | 100 |
| Number of training vectors: | 3000 |
| Number of testing vectors: | 1500 |

Equation (8) is used for the decision layer weights in a network with 0 to 6 pre-decision layers all with random weights. One set of resulting training and testing accuracies are presented in Table 1.

| No. Pre Layers | Train Accuracy | Test Accuracy |
|---|---|---|
| 0 | 0.49 | 0.395 |
| 1 | 0.51 | 0.406 |
| 2 | 0.461 | 0.372 |
| 3 | 0.275 | 0.221 |
| 4 | 0.145 | 0.137 |
| 5 | 0.138 | 0.124 |
| 6 | 0.129 | 0.13 |

Table 1: Accuracies using Computed Decision Weight

Compared to the base accuracy of 0.1, the accuracies produced by computed decision weights are quite good for 0 to 2 layers with random weights, but degrade quickly as more interposing layers with random weights are inserted.

The best case in Table 1 is for a single random pre-decision layer. For that case consider the question: can the same results be learned? For this experiment we keep the random weights in the single pre-decision layer fixed, and starting with random weights, adjust the decision layer weights according the formula

$$\Delta w_i = -\alpha \nabla_{w_i} L = \alpha[M_i - (YY')w_i] \tag{15}$$

For such a simple weight change formula a suitable choice for the learning rate $\alpha$ is surprisingly difficult to find. We tried a number of reasonable values for $\alpha$ and all gave poor and erratic learning performances. Some of the "better" results are shown in Figure 2.

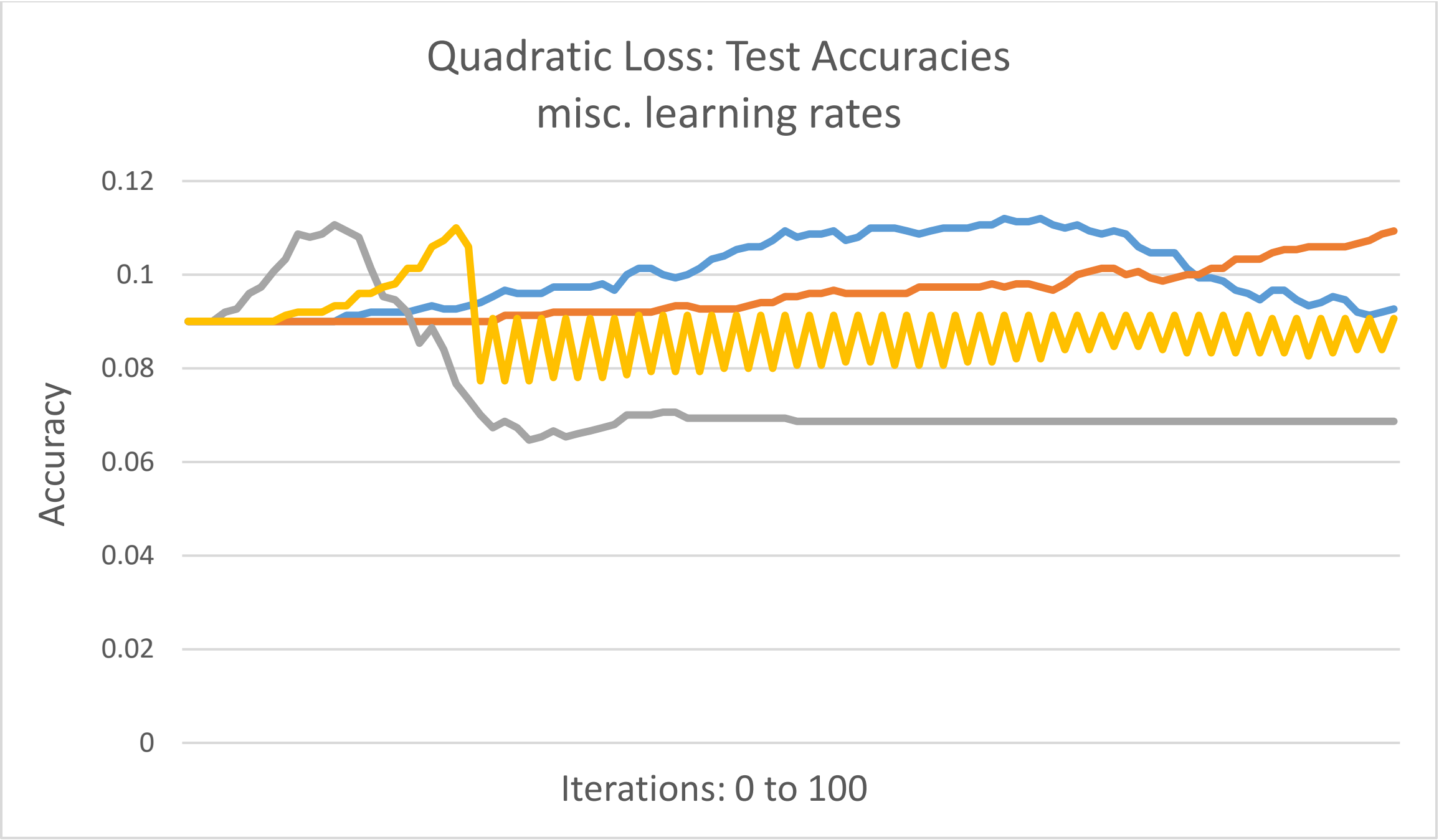

Figure 2. Conventional Gradient Descent

It seems unlikely that the test probability of .406 that results from computed decision weights can be reached by learning through gradient descent. Indeed that is the case as is seen in the next section where the linearity of (15) is used to explore the learning dynamics of gradient descent.

Dynamics of Gradient Descent

Adding $w_i$ to both sides of (15) results in a linear recursion

$$w_i^{(n+1)} = [I - \alpha(YY')]w_i^{(n)} + \alpha M_i \quad (16)$$

where $w_i{}^{(n)}$ denotes the nth iteration of $w_i$ in the learning process. Equation (16) has an explicit solution given by

$$w_i^{(n)} = (YY')^{-1}M_i + [I - \alpha(YY')]^n(w_i^{(0)} - (YY')^{-1}M_i] \quad (17)$$

We can write $YY' = D\Lambda D'$ where $D$ is a unitary matrix and $\Lambda$ is a diagonal matrix with eigenvalues of $YY'$ as its entries. Hence we can write

$$[I - \alpha(YY')]^{(n)} = D[I - \alpha\Lambda]^n D' \quad (18)$$

Therefore, the dynamics of (16) is completely characterized by $(1 - \alpha\lambda)^n$ where $\lambda$ are the eigenvalues of $YY'$. For example, (16) diverges if $\alpha\lambda_{max} > 2$, and oscillates if $1 < \alpha\lambda_{max} < 2$. Similarly, (16) stalls if $\alpha\lambda_{min} \approx 0$.

For (16) to provide effective learning it is reasonable to require $\alpha\lambda_{max} < .25$ (to keep any oscillation to less than 25%) and $\alpha\lambda_{min} > .0025$ (to ensure a minimum change of 25% in 100 iterations). These conditions result in a simple criterion $\left(\frac{\lambda_{max}}{\lambda_{min}}\right) < 500$ on the matrix $YY'$ to achieve effective learning.

For a square matrix its *trace* is defined as the sum of its diagonal elements [Ro 2013]. If we denote the dimension of the data vector $y(x)$ by $N$, then

$$Trace(YY') < N\lambda_{max} \quad \text{and} \quad Trace((YY')^{-1}) < N(\frac{1}{\lambda_{min}})$$

Hence, we have

$$\left(\frac{\lambda_{max}}{\lambda_{min}}\right) > \frac{1}{N^2} Trace(YY')Trace((YY')^{-1}) \quad (19)$$

Trace is relatively easy to compute and we now have a feasible means for assessing the learning efficacy of gradient descent in advance.

For our example of CIFAR-10 dataset and one 100x100 pre-decision layer with random weights, we find $Trace(YY') = 4.13 \times 10^6$, $Trace((YY')^{-1}) = 2.06 \times 10^{11}$ and

$$\left(\frac{\lambda_{max}}{\lambda_{min}}\right) > 8.5 \times 10^{13} \quad (20)$$

No wonder a suitable learning rate is hard to find!

Modifying Gradient Descent

Fortunately, a simple solution is to eliminate the effect of eigenvalues of $YY'$ altogether by modifying the weight change equation to read

$$\Delta w_i = \alpha (YY')^{-1} \nabla_{w_i} L \quad (21)$$

With this change (17) becomes

$$w_i^{(n)} = (YY')^{-1} M_i + (1-\alpha)^n [w_i^{(0)} - (YY')^{-1} M_i] \quad (22)$$

As long as the learning rate stays away from 1 and 0, the dynamics of (22) is as smooth and as predictable as it can be. For example, by setting $\alpha = .5$, we get the learning trajectory shown in Figure 2. We note that the learning rate is chosen to be very aggressive to show stability in even extreme cases. The equilibrium accuracies are reached in 10 iterations. Following a small overshoot at iteration 11, the trajectories stay in steady state thereafter.

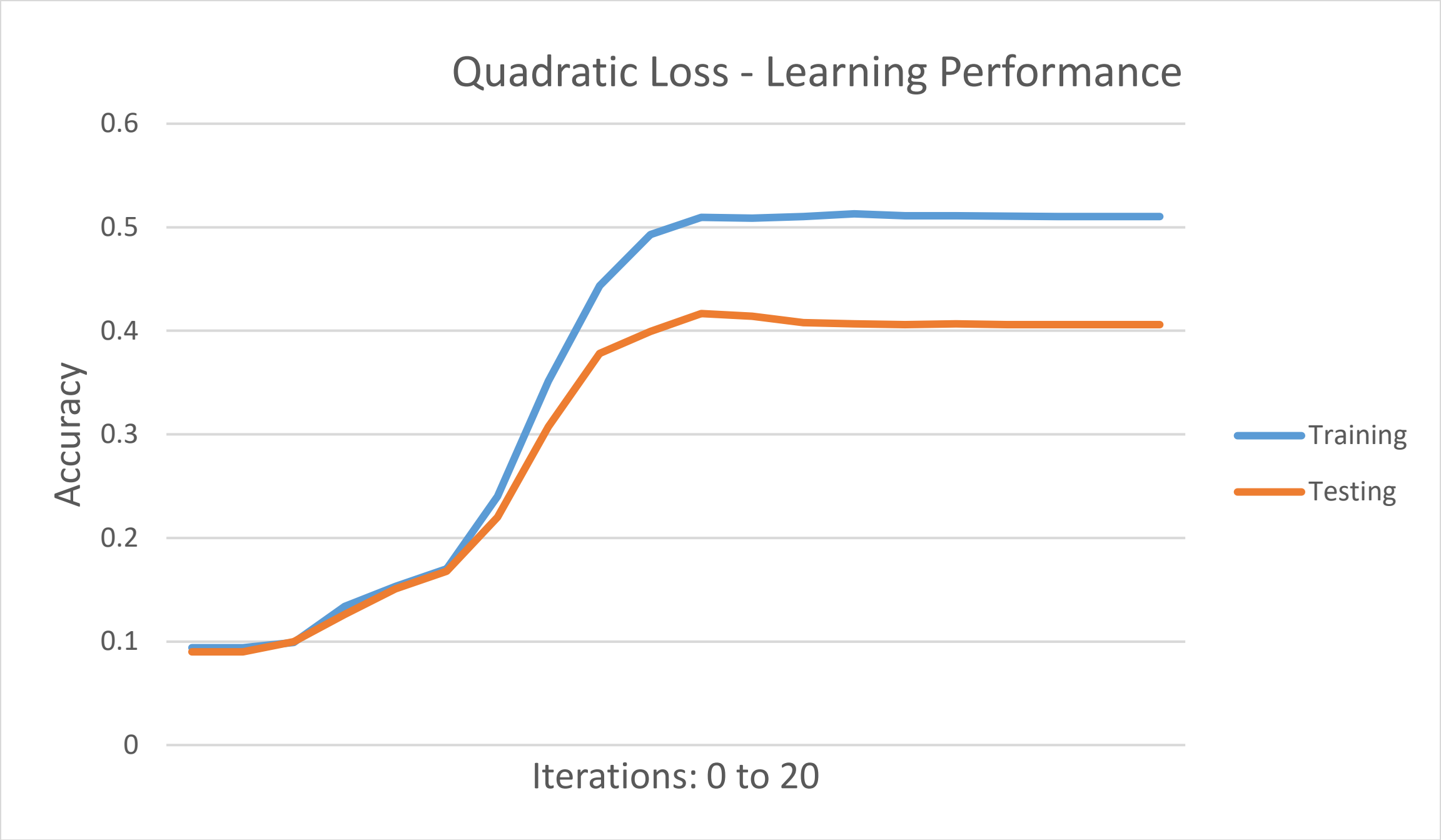

Figure 3. Modified Gradient Descent – Quadratic Loss

A Lagrangian Formulation for the Decision Layer

We note that both cross entropy and quadratic loss try to do the same thing, namely, to increase $z_{c(x)}$ without increasing $z_i$ for $i \neq c(x)$. The same objective can be achieved by solving the following constrained minimization problem using Lagrange multiplier [Be 1982]:

$$\text{Maximize} \quad \sum_{x \in X} z_{c(x)}(x) \qquad \text{subject to} \qquad \sum_{i \leq K, x \in X} z_i^2(x) = \sigma^2 \quad \text{a constant} \tag{23}$$

The maximization in (23) is with respect to the decision weights $w$. To do this we introduce the function

$$L(w, \lambda) = \sum_{x \in X} z_{c(x)} - \lambda\left[\sum_{i \leq K, x \in X} z_i^2(x) - \sigma^2\right] \tag{24}$$

and look for stationary solutions, i.e., those values of $(w, \lambda)$ that satisfy

$$\frac{\partial L}{\partial \lambda} = 0 \qquad \text{and} \qquad \nabla_{w_i} L = 0 \tag{25}$$

From (24) we have

$$\nabla_{w_i} L(w, \lambda) = M_i - 2\lambda(YY')w_i = 0 \tag{26}$$

and

$$\frac{\partial L(w,\lambda)}{\partial \lambda} = \sigma^2 - \sum_{i,x} z_i^2\,(x) = 0 \tag{27}$$

Equation (26) yields

$$w_i = (\tfrac{1}{2\lambda})\rho M_i \tag{28}$$

With

$$\rho = (YY')^{-1} \tag{29}$$

Using (28) in (27), we get

$$\sum_{i,x} z_i^2\,(x) = \sum_i w_i'\,(YY')w_i = (2\lambda)^{-2} \sum_i M_i' \rho M_i = \sigma^2 \tag{30}$$

It follows that

$$\lambda = \left(\frac{1}{2\sigma}\right)\sqrt{\sum_i M_i' \rho M_i} \tag{31}$$

Denoting

$$Z = \sqrt{\sum_i M_i' \rho M_i} \tag{32}$$

we can express the stationary points of $L(w, \lambda)$ as

$$\lambda = Z/2\sigma \qquad \text{and} \qquad w_i = \left(\frac{\sigma}{Z}\right)\rho M_i \tag{33}$$

The solution of the constrained maximum is given by

$$\sum_{x \in X} z_{c(x)} = \sum_{i \le K} w_i' \, M_i = \sigma Z \tag{34}$$

We note that $\sigma$ is a multiplicative constant for both the weights and the objective function. It is merely a scaling factor and we can set $\sigma = 1$.

Aside from $\sigma$ there are no other arbitrary constants. The computed decision weights are precisely $\rho M_i/Z$, not $\rho M_i$ or some constant multiple of it.

We have now formulated the problem of determining the decision weights, not in terms of loss functions or training, but as a problem of constrained maximization. The solution of the problem consists of not only the optimum decision weights, but also the constrained maximum $Z$. They now form the foundation for the problem of learning the pre-decision weights.

Learning Pre-Decision Weights

Freed from the burden of maintaining a constraint, the pre-decision weights can focus on the task of increasing $z_{c(x)}$ as much as possible. To do that we can make weight change proportional to the gradient of the objective function $Z = \sqrt{\sum_i M_i' \rho M_i}$ with respect to the weight vectors of each layer. The basic problem is to compute $\frac{\partial Z}{\partial u_{jk}}$ for any pre-decision weight $u_{jk}$. The dependency of $Z$ on any pre-decision weight goes through the output $y(x)$. The calculation of $\frac{\partial Z}{\partial u_{jk}}$ is usually done in two steps. First, we try to find an expression of the form

$$\frac{\partial Z}{\partial u_{jk}} = \sum_{x \in X} f(x) \frac{\partial y(x)}{\partial u_{jk}}$$

Then back propagation is then used to find $\frac{\partial y(x)}{\partial u_{jk}}$. The derivation for the first part is presented in the Appendix. Equation (A5) of the Appendix yields

$$\frac{\partial Z}{\partial u_{jk}} = z^{-1} \sum_{i \le K, x \in X} [\delta_{ic(x)} - (\rho M_i)' y(x)] (\rho M_i)' \frac{\partial y(x)}{\partial u_{jk}} \tag{35}$$

Linearizing Back Propagation

The nonlinearity in the activation functions is essential to preserve the dimensionality of the weight space. Without it the pre-decision weights collapse into a single layer. On the other hand the nonlinearity in back propagation is needed only to achieve the exact values of the gradients. In certain situations approximate values of the gradients obtained by linearizing back propagation process may suffice. We have explored this possibility and found that for our choice of activation function, hyperbolic tangent, and our data that is indeed the case. Clearly, the specific activation function matters. Data may also matter. But we have no information on that score.

Linearizing back propagation indeed achieves a great deal of simplification in computing the gradients. The notational convention is depicted in the block diagram for a single pre-decision layer shown in Figure 4.

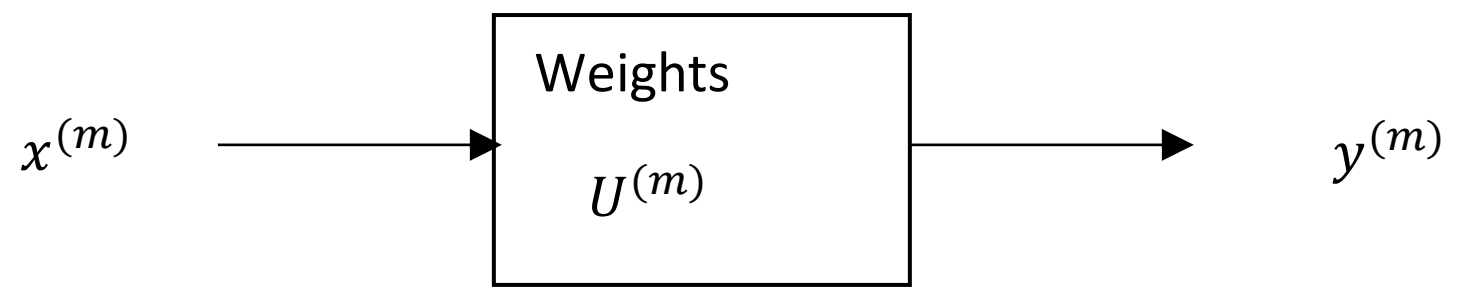

Figure 4. A Single Pre-decision Layer

We designate $N$ as the number for the final layer that precedes the Decision Layer. Assuming a linearization of back propagation, we can compute the gradients for the pre-decision weights in two steps. First we compute a set of equivalent decision weights for each layer using backward recursion.

Equivalent decision weights: $w_i^{(m)} = (U^{(m+1)})' w_i^{(m+1)}$ with $w_i^{(N)} = w_i$ (36)

Second, we compute for each layer a set of input class vectors.

Input class vector for class $i$: $\mu_i^{(m)} = \sum_{x \in X} [\delta_{ic(x)} - z_i(x)] x^{(m)}(x)$ (37)

We can now write the gradient for layer m as

$$\nabla_{U^{(m)}} Z = w^{(m)} (\mu^{(m)})' \qquad (38)$$

where $w^{(m)}$ and $\mu^{(m)}$ are matrices with columns $w_i^{(m)}$ and $\mu_i^{(m)}$ respectively.

Some Experimental Results

We have conducted some experiments on our data set using a classifier with 8 pre-decision layers each with a 100x100 weight matrix. The pre-decision weights are adjusted using the gradient formula given by (38) and we show the increase in accuracy for both training and testing data for a 400 iteration run in Figure 5.

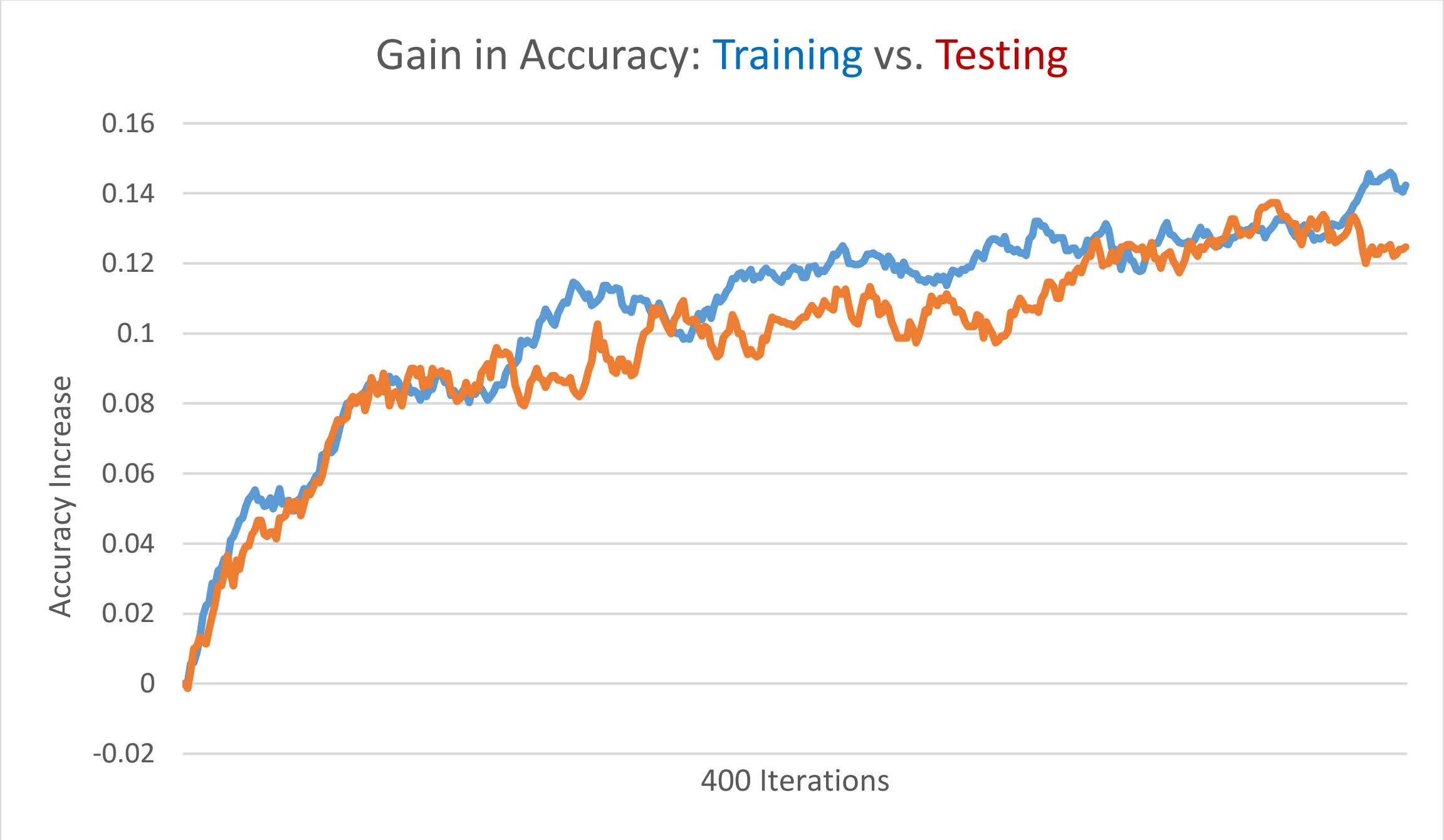

Figure 5. Learning Trajectory – Linearized Back propagation

The learning trajectories depicted in Figure 5 are encouraging for two reasons. First, they show robust progress. Second, they also indicate a good capacity for generalization. The testing accuracy improves almost as well as the training accuracy. There is no evidence of over-fitting the training data.

The combination of computed decision weights together with training pre-decision weights using linearized back propagation also shows promise for deep networks. Experiments with up to eight layers show that this approach is accretive, i.e., more layers the better. Figure 6 shows a comparison between 4 and 8 layers.

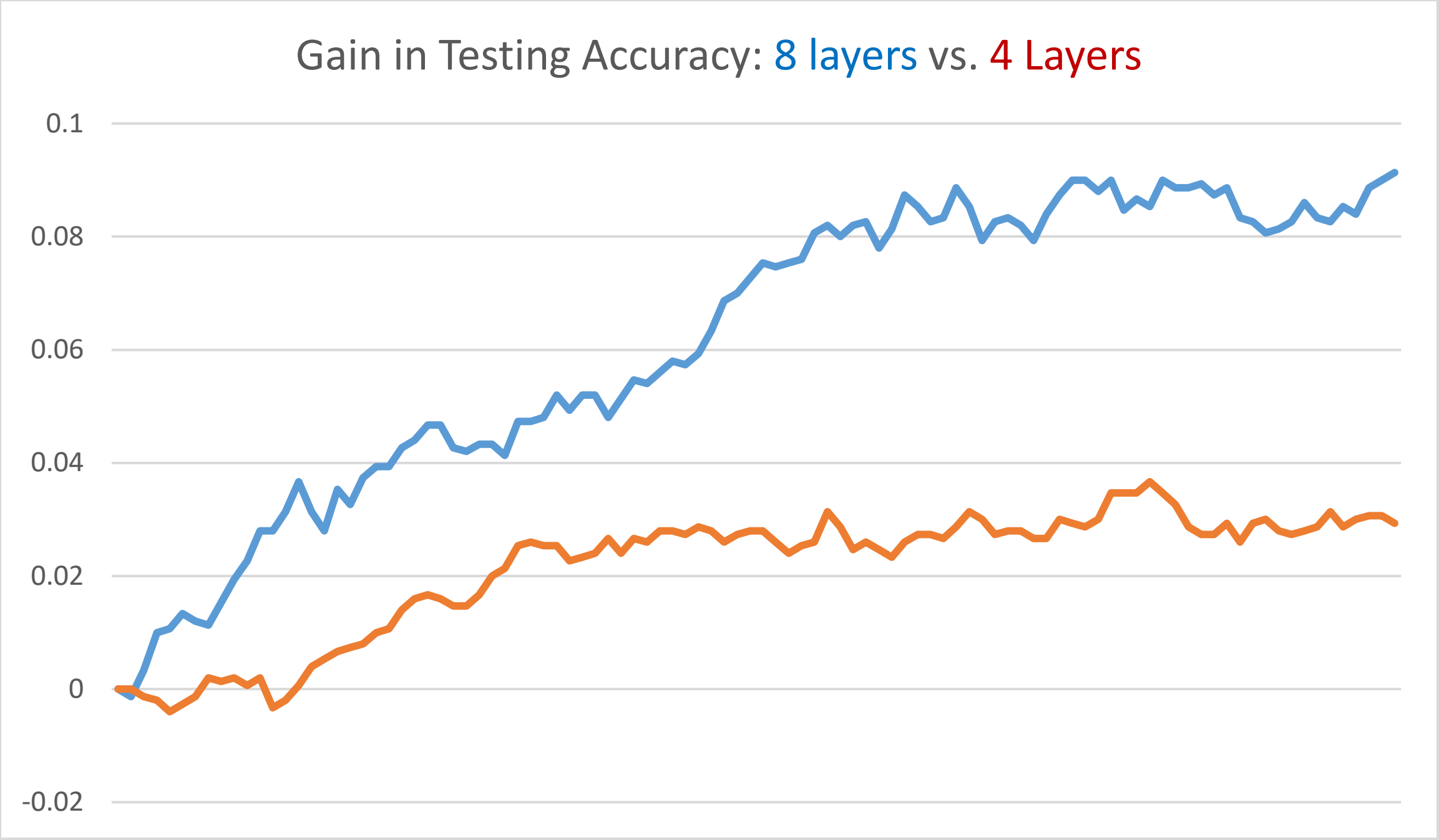

Figure 6. Effect of More Layers

Figure 6 shows that 8 layers are more than twice as good as 4 layers. Although the experimental evidence is limited, there is strong evidence that the new learning algorithm is accretive in the number of layers, possibly even super-accretive for a small number of layers.

A New learning Paradigm for Neural Classifier

The combination of computed decision weights together with a linearization of back propagation leads to a new learning algorithm with considerable promise. The components of the algorithm can be summarized as follows:

Information extracted prior to the decision layer: $M_i = \sum_{x \in X} \delta_{ic(x)} y(x) \quad \rho = (\sum_{x \in X} y(x) y'(x))^{-1}$

Solution to constrained maximization: $Z = \sum_{i \leq K} M_i' \rho M_i \qquad w_i = \rho M_i / Z$

Equivalent decision weights for layer $m$: $w_i^{(m)} = U^{(m+1)} w_i^{(m+1)} \quad w_i^{(N)} = w_i$

Input class vector for layer $m$: $\mu_i^{(m)} = \sum_{x \epsilon X} [\delta_{ic(x)} - z_i(x)] x^{(m)}(x)$

Weight change formula for layer $m$: $\Delta U^{(m)} = \beta w^{(m)} (\mu^{(m)})' \quad \beta =$ positive constant

Summary

In this paper we show that the decision weights of a neural classifier can be computed through the use of a quadratic loss function or as the solution of a constrained optimization problem. The results are similar, but not the same. In the latter case the results include an objective function on which the training of pre-decision weights can be based. We also show that computed decision weights together with a linearization of back propagation lead to a new learning algorithm that show both efficiency and efficacy in limited experimental computation.

Appendix

Here we present a derivation of the formula for $\frac{\partial Z}{\partial u}$, where $u$ is any pre-decision weight and $Z$ is given by (32). We write

$$\frac{\partial Z}{\partial u} = \left(\frac{1}{2}\right) Z^{-\frac{1}{2}} \frac{\partial}{\partial u} \sum_{i \leq K} M_i' \rho M_i$$

$$= \frac{1}{2} Z^{-1/2} \sum_i [(\frac{\partial M_i}{\partial u})' \rho M_i + M_i' \rho \frac{\partial M_i}{\partial u} + M_i' \frac{\partial \rho}{\partial u} M_i] \quad \text{(A1)}$$

The first two terms in the sum on the rights hand side of (A1) are both scalars and mutually transposed. Hence they are equal and their sum can be written as

$$(\frac{\partial M_i}{\partial u})' \rho M_i + M_i' \rho \frac{\partial M_i}{\partial u} = 2 \sum_{x \in X} \delta_{ic(x)} (\rho M_i)' \left(\frac{\partial y(x)}{\partial u}\right) \quad \text{(A2)}$$

The last term involves $\frac{\partial \rho}{\partial u}$. Since $\rho = (YY')^{-1}$, we can write

$$\frac{\partial \rho}{\partial u} (YY') + \rho \frac{\partial}{\partial u} (YY') = 0$$

Hence $$\frac{\partial\rho}{\partial u} = -\rho\frac{\partial}{\partial u}(YY')\rho = -\rho\sum_{x\in X}[\frac{\partial y(x)}{\partial u}y'(x) + y(x)(\frac{\partial y(x)}{\partial u})']\rho \qquad \text{(A3)}$$

The two terms on the right hand side of (A3) are both equal to the inner product of two vectors $\rho y(x)$ and $\rho\frac{\partial y(x)}{\partial u}$. They are equal and it matters not which order the two vectors are written. Hence

$$\frac{\partial\rho}{\partial u} = -2\sum_{x\in X}(\rho y(x))'(\rho\frac{\partial y(x)}{\partial u}) \qquad \text{(A4)}$$

Combining (A2) and (A4) in (A1) we get

$$\frac{\partial Z}{\partial u} = z^{-1}\sum_{i\le K, x\in X}[\delta_{ic(x)} - (\rho M_i)'y(x)](\rho M_i)'\frac{\partial y(x)}{\partial u} \qquad \text{(A5)}$$

**Acknowledgement**

I am grateful to Dr. J. M. Ho of Academia Sinica and Professor C. Y. Lee of the National Chiao Tung University (NCTU) in Taiwan for many helpful comments. Eugene Lee of the NCTU team provided me with the CIFAR-10 data and made many useful comments on a draft of this paper.